\documentclass{article}

\usepackage[final]{neurips_2022}
\usepackage{natbib}

\usepackage[utf8]{inputenc} % allow utf-8 input
\usepackage[T1]{fontenc}    % use 8-bit T1 fonts
\usepackage{hyperref}       % hyperlinks
\usepackage{url}            % simple URL typesetting
\usepackage{booktabs}       % professional-quality tables
\usepackage{amsfonts}       % blackboard math symbols
\usepackage{nicefrac}       % compact symbols for 1/2, etc.
\usepackage{microtype}      % microtypography
\usepackage{xcolor}         % colors
\usepackage{graphicx}
\usepackage{multicol}
\usepackage{multirow}

\title{Spread Love Not Hate: Undermining the Importance of Hateful Pre-training for Hate Speech Detection}

\author{Omkar Gokhale$^1$\thanks{~first author, equal contribution}~~, 
        Aditya Kane$^1$$\footnotemark[1]$~,
        Shantanu Patankar$^1$$\footnotemark[1]$,
        Tanmay Chavan$^1$$\footnotemark[1]$,
        Raviraj Joshi$^2$\\
  Pune Institute of Computer Technology, L3Cube$^1$\\
  Indian Institute of Technology Madras, L3Cube$^2$\\
  \texttt{\{omkargokhale2001}, \texttt{adityakane1},  \texttt{shantanupatankar2001\}@gmail.com},\\ \texttt{\{chavantanmay1402}, \texttt{ravirajoshi\}@gmail.com} \\
}

\begin{document}

\maketitle

\begin{abstract}
%   We introduce four new models Marathi-Tweet-BERT, Marathi Hate BERT ,Marathi non hate BERT and random Marathi 1m BERT. These models were pre-trained on biased subsets of 40M Marathi tweets corpus. We further explore the significance of domain specific pre-training in hate detection in Marathi language. We have tried to answer the question “Does Hateful BERT have leverage in identifying hate?”. We present the results of a detailed comparison of our models with MahaBERTv2 and MuRiL.

Pre-training large neural language models, such as BERT, has led to impressive gains on many natural language processing (NLP) tasks. Although this method has proven to be effective for many domains, it might not always provide desirable benefits. In this paper, we study the effects of hateful pre-training on low-resource hate speech classification tasks. While previous studies on the English language have emphasized its importance, we aim to augment their observations with some non-obvious insights. We evaluate different variations of tweet-based BERT models pre-trained on hateful, non-hateful, and mixed subsets of a 40M tweet dataset. This evaluation is carried out for the Indian languages Hindi and Marathi. This paper is empirical evidence that hateful pre-training is not the best pre-training option for hate speech detection. We show that pre-training on non-hateful text from the target domain provides similar or better results. Further, we introduce HindTweetBERT and MahaTweetBERT, the first publicly available BERT models pre-trained on Hindi and Marathi tweets, respectively. We show that they provide state-of-the-art performance on hate speech classification tasks. We also release hateful BERT for the two languages and a gold hate speech evaluation benchmark HateEval-Hi and HateEval-Mr consisting of manually labeled 2000 tweets each. The models and data are available at https://github.com/l3cube-pune/MarathiNLP .

\end{abstract}

\section{Introduction}
\label{sec:introduction}

Detecting hate speech in social media is a crucial task \citep{schmidt2017survey,velankar2022review}. The effect of hateful social media content on the mental health of society is still under research, but it is undeniably negative \citep{kelly2018social,de2014mental}. Twitter is a powerful social media platform and is quite popular in India for the past few years. It is used by many politicians, activists, journalists, and businessmen as an official medium of communication with society. Though Article 19 in the Indian Constitution guarantees freedom of speech and expression, identifying and curbing hateful tweets is essential to maintain harmony.

Hindi and Marathi are Indo-Aryan languages predominantly spoken in India. 
% Hindi is one of the two official languages in India, while Marathi is the official language of the Indian state of Maharashtra.
Both languages are derived from Sanskrit and have 40+ dialects. Marathi is spoken by 83 million people. It is the third-largest spoken language in India and the tenth in the world. Hindi, being spoken by 528 million people, is the largest spoken language in India and the third largest in the world.

Identification of hate in NLP has followed a common trend in NLP, the manual feature-based classifiers were followed by CNNs and LSTMs, which were then superseded by the modern pre-trained transformers \citep{mullah2021advances,badjatiya2017deep,velankar2022mono}. The transformer-based masked language models (MLM) pre-trained on a variety of text data are suitable for general-purpose use cases.

Creating a domain-specific bias in a pre-training corpus has previously shown state-of-the-art results \citep{gururangan2020don}.  Thus, in this paper, we try to find the impact of hateful pre-training on hate speech classification. The previous work has shown the positive impact of using Hateful BERT for downstream hate speech identification tasks \citep{caselli2020hatebert,sarkar2021fbert}. However, it remains to be verified that the improvements were indeed due to the hateful nature of the pre-training corpus or are simply a side effect of adaptation to target domain text. The past work in the high-resource language is thus incomplete and does not provide sufficient evidence to analyze the impact of hateful pre-training. To complete the analysis, we pre-train our models using both hateful and non-hateful data from the target domain. Moreover, there is no previous work related to hateful pre-training in low-resource Indic languages.  Our work also tries to fill this gap for low-resource languages. 

While evaluating the impact of pre-training, we build some useful resources for Hindi and Marathi. We introduce two new models MahaTweetBERT\footnote{\href{https://huggingface.co/l3cube-pune/marathi-tweets-bert}{MahaTweetBERT}} and HindTweetBERT\footnote{\href{https://huggingface.co/l3cube-pune/hindi-tweets-bert-v2}{HindTweetBERT}} pre-trained on 40 million Marathi and Hindi tweets respectively. We use these models along with MuRIL \cite{khanuja2021muril}, the state-of-the-art Indic multilingual BERT to generate baseline results. To extract the most hateful and least hateful tweets from these 40 million corpora, we classify the tweets using previous state-of-the-art models and choose tweets with the highest confidence (most hateful) and lowest confidence (least hateful). We show that the selected data is indeed hateful by randomly choosing 2000 samples each for both languages and labeling them manually. To actually see if hateful pre-training has an impact, we compare the performances of models pre-trained on the most hateful, least hateful, and random corpora against our baseline on downstream hate speech identification tasks. We show that hateful pre-training is helpful when considered in isolation, however non-hateful or random pre-training is equivalently good. The improvement in performance with hateful pre-training could be a side effect of target domain adaptation and is not dependent upon the hatefulness of the pre-training corpus.  The hateful models are termed as MahaTweetBERT-Hateful \footnote{\href{https://huggingface.co/l3cube-pune/marathi-tweets-bert-hateful}{MahaTweetBERT-Hateful}} and HindTweetBERT-Hateful\footnote{\href{https://huggingface.co/l3cube-pune/hindi-tweets-bert-hateful}{HindTweetBERT-Hateful}}. The 40M tweets corpus is termed as L3Cube-MahaTweetCorpus and HindTweetCorpus for Marathi and Hindi respectively.  The datasets and models released as a part of this will be documented on github\footnote{\href{https://github.com/l3cube-pune/MarathiNLP}{MarathiNLP}} as well.

The main contributions of this work are as follows.
\begin{itemize}
 \item We show that hateful BERT is not always desirable for hate speech detection tasks, and the BERT model pre-trained on non-hateful in-domain data yields similar or better performance.
 \item We release pre-trained Twitter BERT models MahaTweetBERT and HindTweetBERT for Marathi and Hindi. We also release MahaTweetBERT-Hateful and HindTweetBERT-Hateful, the hateful version of the corresponding models. These models are fine-tuned versions of current state-of-the-art MahaBERT and HindBERT models on corresponding language tweets data (40 M sentences).
 \item We release gold standard benchmark hate speech detection datasets HateEval-Mr and HateEval-Hi with 2000 manually labeled tweets.
\end{itemize}

\begin{figure}[]
  \centering
  \includegraphics[height=3.5cm]{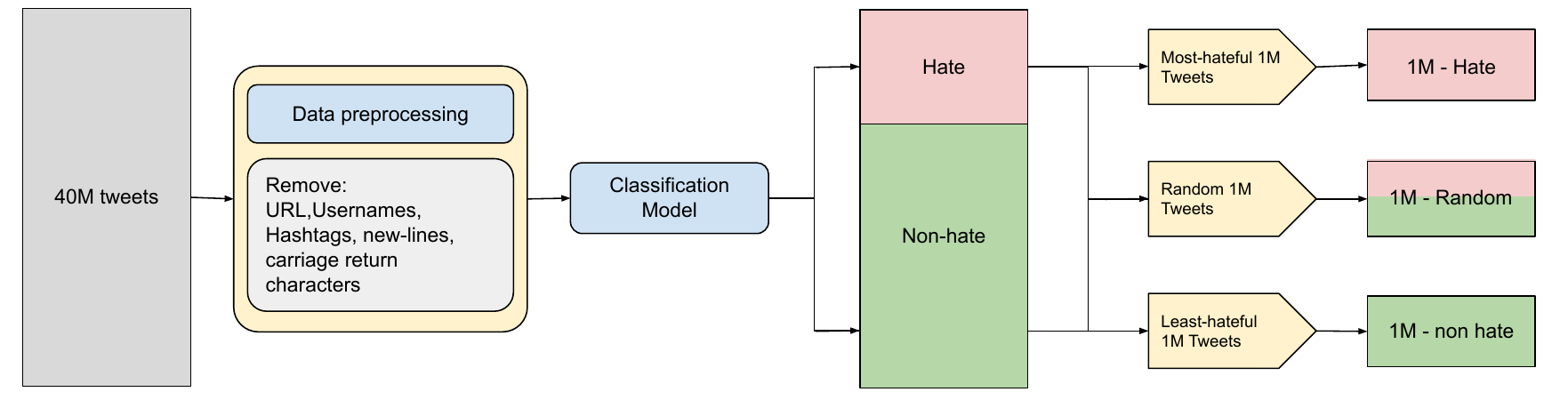}
  \caption{Dataset Generation}
  \label{fig:dataset}
\end{figure}
\section{Related Work}
\label{sec:related_work}

% \begin{figure}[]
% \includegraphics{L3Cube Dataset Creation.pdf}
% \caption{\bf {\large Overview of the integrity discovery system using secure introspection}}
% \end{figure}

% For example, SciBERT \citep{beltagy2019sciBERT} is a model exclusively pretrained on scientific text to improve performance on downstream scientific NLP tasks like sentence classification, sequence tagging, and dependency parsing.

Pre-trained models have obtained remarkable results in many areas of NLP. Although these pre-trained models are well suited for generalized tasks, they have some limitations in domain-specific tasks. To combat this, numerous domain-specific models have been developed based on the BERT  architecture \citep{devlin2018bert}. Domain-specific NLP models are pre-trained on in-domain data that is unique to a specific category of text. For example, BioBERT \citep{lee2020bioBERT} is a model that is trained on large-scale biomedical corpora. It outperforms the previous state-of-the-art models on tasks like biomedical named entity recognition and biomedical question answering. Similarly, ClinicBERT \citep{huang2019clinicalBERT}, FinBERT \citep{yang2020finBERT}, LEGAL-BERT \citep{chalkidis2020legal}, SciBERT \citep{beltagy2019scibert} are models that are pre-trained on clinical notes, financial data, legal documents, and scientific text respectively. They show significantly better performance on downstream tasks in their respective domains.

This concept can also be extrapolated for hate speech detection. Models that are pre-trained on exclusively hateful or non-hateful data could work better than models pre-trained on mixed data. For example, HateBERT \citep{caselli2020hatebert} is a BERT-based model retrained on hateful data extracted from the RAL-E dataset. The RAL-E dataset contains English comments obtained from various subreddits, of which 1.4 million are hateful and are used for pre-training. Similarly, FBERT \citep{sarkar2021fbert} is a BERT-based model trained on 1.4 million exclusively hateful tweets from the SOLID dataset \citep{rosenthal2020solid}. The SOLID dataset contains 9 million English tweets, of which 1.4 million hateful tweets are used for pre-training. Both these models work better than a vanilla BERT model, when fine-tuned on the downstream training data. Though the results obtained by HateBERT and F-BERT are valid, their experimental setup is not exhaustive and lacks crucial ablations. Given this, we perform a systematic and more exhaustive study of pre-training on Hindi and Marathi data. We test this on two languages to ensure that the results are not language-specific.

There are various models used for hate speech detection in Marathi. These include monolingual models like MahaBERT \citep{joshi2022l3cube} and multilingual models like MuRIL \citep{khanuja2021muril}. Similarly, in Hindi, models like HindBERT \citep{joshi2022l3cubehind}, and MuRIL are the state-of-the-art models for detecting hate speech. We propose a more comprehensive approach than HateBERT and F-BERT, where we pre-train our model on hateful, non-hateful, and random Marathi and Hindi tweets. We then compare the performance of the Marathi models with MuRIL and MahaBERT and the Hindi models with MuRIL and HindBERT to check whether selective pre-training affects model performance.

\section{Dataset description}
\label{sec:dataset}

We create new datasets to observe the effects of using deep learning models trained on primarily hateful data. To test the veracity of our results we have performed experiments with datasets of two different Indian languages, Marathi and Hindi. We follow the same practices and procedures for both languages. The datasets consist of a large number of primarily monolingual tweets. From the obtained corpora, we use a threshold value to ensure that tweets contain the majority of words from the desired language. This ensures that the datasets do not contain tweets with only a few words from the desired language as well as includes tweets that have a small number of words from other languages (primarily common English terms and acronyms such as GST, and CAA). All the tweets from both the corpora contain Devanagiri script characters along with numbers. The tweets originally contained usernames, URLs, hashtags, and emojis. We clean the data and redact usernames, URLs, and hashtags from the datasets to prevent user identification and noise. However, we retain emojis as they might contribute to the semantics of the sentence. All of the tweets are pre-processed before being used for pre-training the models.

\subsection{Pre-training corpus}
We create MahaTweetCorpus and HindTweetCorpus with roughly 40 million tweets each for Marathi and Hindi languages respectively. All of the other datasets are subsets of these corpora. We illustrate our dataset sampling process in Figure \ref{fig:dataset}. We extract four datasets from the 40M tweet corpus.
The primary dataset contains \textbf{all of the scraped tweets}. \\
The second type of dataset contains \textbf{1 million tweets randomly sampled} from the primary dataset. The tweets from the 40M datasets were classified into hateful and non-hateful tweets using existing hate speech classification models in the respective languages. For Marathi, the tweets were classified by using the MahaHateBERT model\footnote{\href{https://huggingface.co/l3cube-pune/mahahate-bert}{MahaHateBERT}}. MahaHateBERT is a variant of BERT fine-tuned on the MahaHate dataset \citep{mahahate}. For Hindi, we use a RoBERTa model \citep{liu2019roberta} fine-tuned on the Hindi dataset for hate speech classification\citep{hasocbert}. We record the prediction confidence values along with the predicted labels. \\
Our third dataset contains \textbf{1 million of the most hateful tweets} based on the confidence values of the model. Our fourth dataset consists of the \textbf{1 million least hateful tweets} as per the confidence values. We choose to use 1 million tweets to maintain consistency with other experiments and also in line with previous work in this area.

\subsection{Pre-training corpus verification}
To ensure that the models used for segregating hateful and non-hateful tweets are reliable, we manually annotate 2000 tweets of each language and compare them with the predicted labels. We see that the classification models, namely MahaBERT and HASOC-RoBERTa, perform reliably. Concretely, they exhibit a classification accuracy of 77\% for Marathi and 75\% for Hindi. Since we sample the most hateful and least hateful data for the respective pre-training tasks, we also calculate the classification model accuracy for the most and least hateful predictions in this verification dataset. We report even better accuracies in this high confidence set. For Marathi, we get 100\% and 96.27\% accuracy for the least and most hateful samples, respectively.
Similarly, in Hindi, we get 95.22\% and 80.32\% accuracy for the least and most hateful samples, respectively. The approach can therefore be seen as a credible data selection strategy for pre-training. After manual annotation, we see that out of 2000 randomly sampled tweets for each language, 803 tweets are hateful in Hindi, and 739 tweets are hateful in Marathi. We plan to release the full verification dataset as a gold benchmark test set HateEval-Hi and HateEval-Mr in near future.
% As we select tweets with the highest confidence, these numbers are merely lower bounds for the actual classification accuracy.

\subsection{Downstream evaluation}
We use three datasets for Marathi and two for Hindi to validate and compare our models.

The \textbf{HASOC 2021 Marathi dataset} is a dataset presented by the Hate Speech and Offensive Content Identification in English and Indo-Aryan Languages (HASOC 2021) track \citep{https://doi.org/10.48550/arxiv.2112.09301} in the Forum for Information Retrieval Evaluation (FIRE 2021). It contains a total of 2,499 tweets in Marathi manually annotated by native speakers of the language. The dataset was used for a shared task organized by HASOC. The tweets are classified into two categories as hateful and non-hateful tweets.

The \textbf{MahaHate two-class dataset} is a hate-speech dataset in Marathi. It contains tweets written in Marathi, annotated manually by four annotators who were fluent in the language. The tweets are classified into two categories, namely hateful or offensive and non-hateful tweets. It contains an equal number of hateful and non-hateful tweets, with a total of 37,500 tweets. The results presented in this paper are based on the models being trained on only 50 percent of the 30,000 tweets of the training dataset. This was primarily done because using the full training data gave a similar performance with all the models and the improvements were more prominent in low-resource settings.
%% Check whether to say that the full dataset gave suboptimal results

The \textbf{MahaHate four-class dataset} contains Marathi tweets and is similar to the above dataset but contains four classes. The tweets are classified into four categories: hate (HATE), offensive (OFFN), profanity (PRFN), and non-hateful (NOT). The dataset contains a total of 25000 tweets, including the train, test, and validation datasets. There is an equal number of tweets in each of the four categories. Similar to the 2-class dataset, we have used only the random half of the training dataset for fine-tuning our models. 

We use two datasets for validating and benchmarking our Hindi models. The \textbf{HASOC 2021 2-class dataset} contains about 4500 tweets. The classes in the two-class dataset are similar to the HASOC 2021 Marathi dataset.

We also use the \textbf{CONSTRAINT 2021 dataset} \citep{https://doi.org/10.48550/arxiv.2011.03588}. The original dataset is designed for multilabel classification. We modify the dataset into a 4-class dataset by removing samples that are irrelevant or have multiple ground truth classes. Our modified dataset contains the classes not offensive (NOT), hateful (HATE), defamation (DEF), and offensive (OFF).

\begin{figure*}[]
    \centering
    \includegraphics[height=4.5cm]{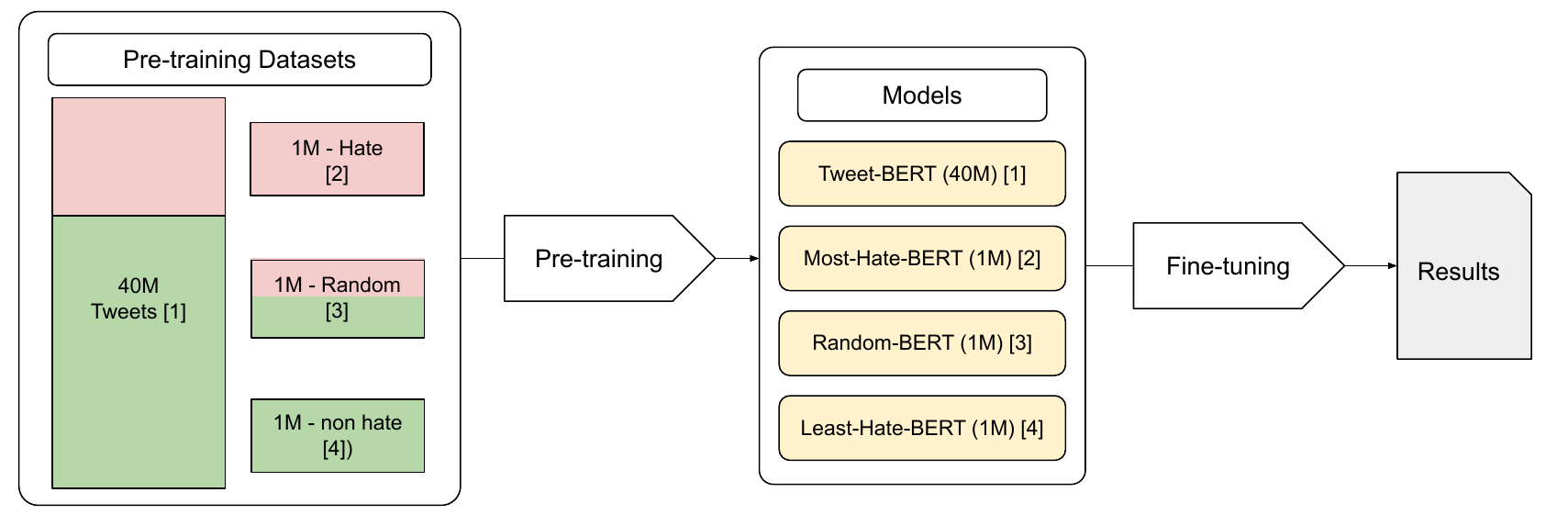}
    \caption{Experimental Setup}
    \label{fig:training}
\end{figure*}

\section{Experiments}
\label{sec:experiments}

We conduct extensive experiments to study the effect of different pre-training data compositions on downstream performance. Specifically, we evaluate the different models on three hate detection datasets - MahaHate 4-class, MahaHate 2-class, and HASOC 2-class for Marathi. Similarly, we use two Hindi versions of datasets to test our Hindi models - CONSTRAINT 4-class and HASOC 2-class. The number of samples used for training, validation, and testing are given in Table \ref{tab:downstream_ds_stats}. 
% Our main focus during experiments has been training models on different subsets of the bigger corpus. 

% Moreover, they completely ignore that a mixed dataset, consisting of hateful and non-hateful articles can be used for pretraining. 

\begin{table*}[]
\begin{tabular}{|ccccc|}
\hline
\multicolumn{1}{|c|}{\textbf{Dataset}}                             & \multicolumn{1}{c|}{\textbf{Model}}         & \multicolumn{1}{c|}{\textbf{Accuracy}} & \multicolumn{1}{c|}{\textbf{F1 macro}} & \textbf{F1 Weighted} \\ \hline
\multicolumn{5}{|c|}{\textbf{Marathi}}                                                                                                                                                                                    \\ \hline
\multicolumn{1}{|c|}{\multirow{6}{*}{\textbf{MahaHate - 2 class}}} & \multicolumn{1}{c|}{MuRIL}         & \multicolumn{1}{c|}{88.57}             & \multicolumn{1}{c|}{88.57}             & 88.57                \\ \cline{2-5} 
\multicolumn{1}{|c|}{}                                             & \multicolumn{1}{c|}{MahaBERT}            & \multicolumn{1}{c|}{89.75}             & \multicolumn{1}{c|}{89.75}             & 89.75                \\ \cline{2-5} 
\multicolumn{1}{|c|}{}                                             & \multicolumn{1}{c|}{MahaTweetBERT}        & \multicolumn{1}{c|}{\textbf{89.94}}             & \multicolumn{1}{c|}{\textbf{89.93}}             & \textbf{89.93}                \\ \cline{2-5} 
\multicolumn{1}{|c|}{}                                             & \multicolumn{1}{c|}{MahaTweetBERT-Hateful} & \multicolumn{1}{c|}{89.47}             & \multicolumn{1}{c|}{89.47}             & 89.47                \\ \cline{2-5} 
\multicolumn{1}{|c|}{}                                             & \multicolumn{1}{c|}{mr-random-twt-1m}       & \multicolumn{1}{c|}{89.52}             & \multicolumn{1}{c|}{89.52}             & 89.52                \\ \cline{2-5} 
\multicolumn{1}{|c|}{}                                             & \multicolumn{1}{c|}{mr-least-ht-1m}         & \multicolumn{1}{c|}{89.57}             & \multicolumn{1}{c|}{89.57}             & 89.57                \\ \specialrule{.15em}{0em}{0em}
\multicolumn{1}{|c|}{\multirow{6}{*}{\textbf{MahaHate - 4 class}}} & \multicolumn{1}{c|}{MuRIL}                  & \multicolumn{1}{c|}{77.83}             & \multicolumn{1}{c|}{77.85}             & 77.85                \\ \cline{2-5} 
\multicolumn{1}{|c|}{}                                             & \multicolumn{1}{c|}{MahaBERT}            & \multicolumn{1}{c|}{79.55}             & \multicolumn{1}{c|}{79.55}             & 79.55                \\ \cline{2-5} 
\multicolumn{1}{|c|}{}                                             & \multicolumn{1}{c|}{MahaTweetBERT}        & \multicolumn{1}{c|}{\textbf{79.7}}              & \multicolumn{1}{c|}{\textbf{79.71}}             & \textbf{79.71}                \\ \cline{2-5} 
\multicolumn{1}{|c|}{}                                             & \multicolumn{1}{c|}{MahaTweetBERT-Hateful}          & \multicolumn{1}{c|}{78.43}             & \multicolumn{1}{c|}{78.49}             & 78.49                \\ \cline{2-5} 
\multicolumn{1}{|c|}{}                                             & \multicolumn{1}{c|}{mr-random-twt-1m}       & \multicolumn{1}{c|}{79.08}             & \multicolumn{1}{c|}{79.15}             & 79.15                \\ \cline{2-5} 
\multicolumn{1}{|c|}{}                                             & \multicolumn{1}{c|}{mr-least-ht-1m}         & \multicolumn{1}{c|}{78.8}              & \multicolumn{1}{c|}{78.88}             & 78.88                \\ \specialrule{.15em}{0em}{0em}
\multicolumn{1}{|c|}{\multirow{6}{*}{\textbf{HASOC}}}              & \multicolumn{1}{c|}{MuRIL}                  & \multicolumn{1}{c|}{85.7}              & \multicolumn{1}{c|}{84.18}             & 85.58                \\ \cline{2-5} 
\multicolumn{1}{|c|}{}                                             & \multicolumn{1}{c|}{MahaBERT}            & \multicolumn{1}{c|}{88.1}              & \multicolumn{1}{c|}{86.76}             & 88.18                \\ \cline{2-5} 
\multicolumn{1}{|c|}{}                                             & \multicolumn{1}{c|}{MahaTweetBERT}        & \multicolumn{1}{c|}{\textbf{89.09}}             & \multicolumn{1}{c|}{\textbf{87.63}}             & \textbf{89.06}                \\ \cline{2-5} 
\multicolumn{1}{|c|}{}                                             & \multicolumn{1}{c|}{MahaTweetBERT-Hateful}          & \multicolumn{1}{c|}{88.96}             & \multicolumn{1}{c|}{87.53}             & 88.95                \\ \cline{2-5} 
\multicolumn{1}{|c|}{}                                             & \multicolumn{1}{c|}{mr-random-twt-1m}       & \multicolumn{1}{c|}{88.13}             & \multicolumn{1}{c|}{86.71}             & 88.19                \\ \cline{2-5} 
\multicolumn{1}{|c|}{}                                             & \multicolumn{1}{c|}{mr-least-ht-1m}         & \multicolumn{1}{c|}{87.9}              & \multicolumn{1}{c|}{86.52}             & 87.98                \\ \hline
\multicolumn{5}{|c|}{\textbf{Hindi}}                                                                                                                                                                                      \\ \hline
\multicolumn{1}{|c|}{\multirow{6}{*}{\textbf{HASOC 2-class}}}      & \multicolumn{1}{c|}{MuRIL}       & \multicolumn{1}{c|}{78.59}             & \multicolumn{1}{c|}{73.39}            & 77.40               \\ \cline{2-5} 
\multicolumn{1}{|c|}{}                                             & \multicolumn{1}{c|}{HindBERT}          & \multicolumn{1}{c|}{80.09}             & \multicolumn{1}{c|}{75.99}            & 79.37               \\ \cline{2-5} 
\multicolumn{1}{|c|}{}                                             & \multicolumn{1}{c|}{HindTweetBERT}        & \multicolumn{1}{c|}{\textbf{80.97}}            & \multicolumn{1}{c|}{\textbf{77.23}}            & \textbf{80.37}               \\ \cline{2-5} 
\multicolumn{1}{|c|}{}                                             & \multicolumn{1}{c|}{HindTweetBERT-Hateful}          & \multicolumn{1}{c|}{78.36}             & \multicolumn{1}{c|}{73.98}             & 77.59                \\ \cline{2-5} 
\multicolumn{1}{|c|}{}                                             & \multicolumn{1}{c|}{hi-random-twt-1m}       & \multicolumn{1}{c|}{79.73}             & \multicolumn{1}{c|}{75.39}             & 78.91                \\ \cline{2-5} 
\multicolumn{1}{|c|}{}                                             & \multicolumn{1}{c|}{hi-least-ht-1m}         & \multicolumn{1}{c|}{79.44}             & \multicolumn{1}{c|}{74.60}            & 78.38                \\ \specialrule{.15em}{0em}{0em}
\multicolumn{1}{|c|}{\multirow{6}{*}{\textbf{CONSTRAINT 4-class}}} & \multicolumn{1}{c|}{MuRIL}       & \multicolumn{1}{c|}{79.39}             & \multicolumn{1}{c|}{36.85}            & 74.46              \\ \cline{2-5} 
\multicolumn{1}{|c|}{}                                             & \multicolumn{1}{c|}{HindBERT}          & \multicolumn{1}{c|}{81.33}             & \multicolumn{1}{c|}{46.49}            & 78.38                \\ \cline{2-5} 
\multicolumn{1}{|c|}{}                                             & \multicolumn{1}{c|}{HindTweetBERT}        & \multicolumn{1}{c|}{\textbf{81.71}}            & \multicolumn{1}{c|}{\textbf{50.80}}            & \textbf{79.86}               \\ \cline{2-5} 
\multicolumn{1}{|c|}{}                                             & \multicolumn{1}{c|}{HindTweetBERT-Hateful}          & \multicolumn{1}{c|}{79.74}            & \multicolumn{1}{c|}{45.18}            & 77.48                \\ \cline{2-5} 
\multicolumn{1}{|c|}{}                                             & \multicolumn{1}{c|}{hi-random-twt-1m}       & \multicolumn{1}{c|}{80.92}             & \multicolumn{1}{c|}{46.12}            & 78.16               \\ \cline{2-5} 
\multicolumn{1}{|c|}{}                                             & \multicolumn{1}{c|}{hi-least-ht-1m}         & \multicolumn{1}{c|}{81.55}            & \multicolumn{1}{c|}{48.81}             & 79.25                \\ \hline
\end{tabular}
 \caption{Finetuning results on downstream datasets using our models. We report the average scores across five runs with different seed values. The best metrics are shown in bold. Note, MahaTweetBERT-Hateful is also referred as MahaTweetBERT-Hateful and HindTweetBERT-Hateful is same as HindTweetBERT-Hateful.}
    \label{tab:final_results}
\end{table*}

As mentioned in Section \ref{sec:related_work}, \citet{sarkar2021fbert} and \citet{caselli2020hatebert} present the comparison between retrained transformers and out-of-the-box pre-trained transformers. However, this comparison is substantially inadequate as it does not consider other data compositions for pre-training. Therefore, our ablations include a comprehensive study of models trained on different subsets of the data. As mentioned in Section \ref{sec:dataset}, we experiment with four subsets of the data:

\begin{itemize}
    \item The complete corpus of 40 million tweets (MahaTweetBERT / HindTweetBERT)
    \item Most hateful 1 million tweets (MahaTweetBERT-Hateful / HindTweetBERT-Hateful)
    \item Least hateful 1 million tweets (mr-least-ht-1m / hi-least-ht-1m)
    \item Randomly sampled 1 million tweets (mr-random-twt-1m / hi-random-twt-1m)
\end{itemize}

We train four models (names as in parenthesis), each corresponding to one subset of the data for each language. We also provide scores on  MuRIL \citep{khanuja2021muril} and MahaBERT \citep{mahaBERT}, which we use as baselines of our experiments. In the case of Hindi, we use two models, namely HindBERT \citep{joshi2022l3cubehind} and MuRIL, as baselines. The MahaBERT and HindBERT are monolingual BERT models pretrained on publicly available Marathi and Hindi datasets, respectively.

For each subset of the data, we further pre-train the MahaBERT/HindBERT using Masked Language Modelling (MLM) objective on the subset of tweets. After the pre-training, we individually fine-tune each model on the downstream datasets as shown in Appendix \ref{app:dataset_statistics}. We finally report the metrics on the test sets of these datasets. Note that we use the same BERT model for all subsets to ensure fairness. We conduct five runs with different seed values for each downstream fine-tuning experiment and report the mean in all cases. Our evaluation pipeline is illustrated in Figure \ref{fig:training}. We report our final results in Table \ref{tab:final_results}.

% and Table \ref{tab:final_results_hi}. 

We use the following hyperparameters in our experiments. For MLM training, we train the models for two epochs at a learning rate of $2e-5$, with a weight decay of $0.01$ and a mask probability of $0.15$. For fine-tuning, we train the models for 25 epochs with a learning rate of $5e-6$ and no weight decay. We use the Hugging Face library \citep{wolf-etal-2020-transformers} for training and hosting our models. All models are uploaded to the HuggingFace platform, and their links are available in Appendix \ref{app:model_links}.

\section{Results}
% \specialrule{.15em}{0em}{0em}

We hereby analyze the results shown in Table \ref{tab:final_results}. We make some key observations regarding the results as follows.

% \begin{enumerate}
%     \item \textbf{Model pre-trained on 40 million dataset performs the best on all downstream tasks:} MahaTweetBERT outperforms all other subsets on all downstream tasks. This result, although is quite expected, ensures that the large-scale pretraining has indeed helped the model. This gives a strong benchmark for future work.
    
%     \item \textbf{Monolingual model outperforms multilingual models:} We observe that MuRIL, the multilingual model trained on 17 Indian languages, has consistently under performed on all datasets. We speculate this is because our monolingual models understand the Marathi language far better than the multilingual model. It is possible that the acquired knowledge from other languages does not aid performance in the case of MuRIL.
    
%     \item \textbf{Hateful, non-hateful or random pretraining does not affect performance (CHECK AGAIN): } Contrary to our intuition, pretraining only on hateful or non-hateful tweets does not improve performance. In fact, of the three subsets (hateful, non-hateful and random), the model trained on random subset tends to perform better on two of three downstream datasets in Marathi and dash of dash downstream datasets in Hindi. 
%     In the case of HASOC, we observe the hateful model to perform better compared to other subsets. However, in the case of MahaHate - 2 class and MahaHate 4 - class, the random model performs better than the other two.
% \end{enumerate}

\begin{enumerate}
    \item \textbf{Hateful BERT is not the best model:} Contrary to our intuition, pre-training only on hateful tweets does not give the best results. In fact, of the three subsets (hateful, non-hateful, and random), the model trained on the random subset tends to perform better on two of three downstream datasets in Marathi and both downstream datasets in Hindi in terms of macro-F1. Moreover, a model trained on non-hateful data performs better than the model trained on hateful content for the majority of the tasks. The hateful models perform better than the baseline MuRIL model, which is in line with the previous works. We suggest that hateful pre-training is helpful over the raw pre-trained models; however, these are not the best alternatives.
    \item \textbf{Monolingual retraining shows improvement over multilingual models:} We observe that MuRIL, the multilingual model trained on 17 Indian languages and billions of tokens, has consistently underperformed on all datasets. On the other hand, we observe that our models retrained on Hindi tweets outperform MuRIL by a large margin. Note that our Hindi models are essentially MuRIL models retrained on Hindi tweets corpus. The same trend can be seen for Marathi as well. We speculate this is because retraining on sizeable corpora of a particular language augments the multi-lingual pre-training. Specifically, in our case, retraining on languages having medium-sized corpora can outperform cumulative semantic knowledge gained from training on multiple large-sized corpora of different languages, as is in the case of MuRIL.
    \item \textbf{Model pre-trained on 40 million datasets performs the best on all downstream tasks:} The models MahaTweetBERT and HindTweetBERT outperform all other models on all downstream datasets. This result, although expected, ensures that large-scale pretraining has indeed helped the model. The model also performs well on other downstream tasks like Sentiment Analysis as shown in Appendix \ref{app:mahasent}. This provides a strong benchmark for future work.
\end{enumerate}

\section{Conclusion}
In this paper, we test the effect of hateful pre-training on hate speech classification. We pre-train two models, MahaTweetBERT and  HindTweetBERT, on 40 million tweets. Additionally, to empirically validate the usefulness of hateful pre-training, we have pre-trained three models on 1 million random, hateful and non-hateful tweets extracted from the aforementioned 40 million tweets for both Marathi and Hindi languages. We compare the performance of these models on standard hate speech detection datasets like HASOC and CONSTRAINT. Our experiments indicate that hateful or non-hateful pre-training does not define the model performance, as we observe that models pre-trained on biased tweets do not outperform the ones pre-trained on random tweets. We also observe that our monolingual models fair better than multilingual models like MuRIL. Furthermore, we observe that the models pre-trained on all 40 million tweets perform better than the other, relatively smaller, models. These results are consistent for both Marathi and Hindi languages ensuring that these observations are not language-specific.

% In this paper we have presented two new models, marathi-tweets-bert and hindi-tweets-bert. To empirically validate the usefulness of hateful pre-training, we have pre-trained 3 models on 1 million random, hateful and non-hateful tweets for both Marathi and Hindi languages. Our experiments demonstrate that biased pre-training is not effective for detecting hate speech. We also observed that although biased pre-training isn't effective for hate speech detection, the the size of pre-training corpus has high correlation with the performance of these BERT based models.

\section*{Acknowledgements}
This work was done under the L3Cube Pune mentorship
program and a part of the L3Cube-MahaNLP project \cite{joshi2022l3cube}. We would like to express our gratitude towards
our mentors at L3Cube for their continuous support and
encouragement.

\bibliography{anthology, custom}
\bibliographystyle{acl_natbib.bst}

\appendix

\section{Dataset statistics}
\label{app:dataset_statistics}
\begin{table}[h]
\centering
\begin{tabular}{|c|ccc|cc|}
\hline
Split      & \multicolumn{3}{c|}{Marathi}                                    & \multicolumn{2}{c|}{Hindi}              \\ \hline
           & \multicolumn{1}{c|}{HASOC} & \multicolumn{2}{c|}{MahaHate}      & \multicolumn{1}{c|}{HASOC} & CONSTRAINT \\ \hline
 & \multicolumn{1}{c|}{2-class} & \multicolumn{1}{c|}{2-class} & 4-class & \multicolumn{1}{c|}{2-class} & 4-class \\ \hline
Train      & \multicolumn{1}{c|}{1667}  & \multicolumn{1}{c|}{15000} & 10750 & \multicolumn{1}{c|}{3675}  & 4238       \\ \hline
Test       & \multicolumn{1}{c|}{625}   & \multicolumn{1}{c|}{3750}  & 2000  & \multicolumn{1}{c|}{919}   & 1217       \\ \hline
Validation & \multicolumn{1}{c|}{207}   & \multicolumn{1}{c|}{3750}  & 1500  & \multicolumn{1}{c|}{1532}  & 603        \\ \hline
\end{tabular}
\caption{Datasets used for training and validation and their respective splits}
\label{tab:downstream_ds_stats}
\end{table}

\section{Model links}
\label{app:model_links}

We present HTTPS URLs for all pre-trained models used in our experiments in Table \ref{tab:model_links}.

\begin{table*}[h]
\centering
\begin{tabular}{|cl|}
\hline
\multicolumn{2}{|l|}{\textbf{Marathi}}                                                                     \\ \hline
\multicolumn{1}{|l|}{\textbf{Model alias}} & \textbf{HTTPS link}                                           \\ \hline
\multicolumn{1}{|c|}{MuRIL}                & \href{https://huggingface.co/google/muril-base-cased}{google/muril-base-cased}   \\ \hline
\multicolumn{1}{|c|}{MahaBERT}          & \href{https://huggingface.co/l3cube-pune/marathi-bert-v2}{marathi-bert-v2}   \\ \hline
\multicolumn{1}{|c|}{MahaTweetBERT}      & \href{https://huggingface.co/l3cube-pune/marathi-tweets-bert}{marathi-tweets-bert}  \\ \hline
\multicolumn{1}{|c|}{MahaTweetBERT-Hateful}        & \href{https://huggingface.co/l3cube-pune/marathi-tweets-bert-hateful}{MahaTweetBERT-Hateful}   \\ \hline
\multicolumn{1}{|c|}{mr-random-twt-1m}     & \href{https://huggingface.co/l3cube-pune/mr-random-twt-1m}{mr-random-twt-1m}   \\ \hline
\multicolumn{1}{|c|}{mr-least-ht-1m}       & \href{https://huggingface.co/l3cube-pune/mr-least-ht-1m}{mr-least-ht-1m}   \\ \hline
\multicolumn{2}{|l|}{\textbf{Hindi}}                                                                       \\ \hline
\multicolumn{1}{|l|}{\textbf{Model alias}} & \textbf{HTTPS link}                                           \\ \hline
\multicolumn{1}{|c|}{MuRIL}                & \href{https://huggingface.co/google/muril-base-cased}{google/muril-base-cased} \\ \hline
\multicolumn{1}{|c|}{HindBERT}        & \href{https://huggingface.co/l3cube-pune/hindi-bert-v2}{hindi-bert-v2} \\ \hline
\multicolumn{1}{|c|}{HindTweetBERT}      & \href{https://huggingface.co/l3cube-pune/hindi-tweets-bert-v2}{hindi-tweets-bert-v2} \\ \hline
\multicolumn{1}{|c|}{HindTweetBERT-Hateful}        & \href{https://huggingface.co/l3cube-pune/hindi-tweets-bert-hateful}{HindTweetBERT-Hateful} \\ \hline
\multicolumn{1}{|c|}{hi-random-twt-1m}     & \href{https://huggingface.co/l3cube-pune/hi-random-twt-1m}{hi-random-twt-1m} \\ \hline
\multicolumn{1}{|c|}{hi-least-ht-1m}       & \href{https://huggingface.co/l3cube-pune/hi-least-ht-1m}{hi-least-ht-1m} \\ \hline
\end{tabular}

\caption{HTTPS links to all pre-trained models}
\label{tab:model_links}
\end{table*}

\section{MahaSent results}
\label{app:mahasent}
\begin{table*}[h]
\centering
\begin{tabular}{|c|c|c|c|}
\hline
\textbf{Model}            & \textbf{Accuracy} & \textbf{F1 macro} & \textbf{F1 Weighted} \\ \hline
MuRiL                     & 0.843                                     & 0.843                                     & 0.843                                        \\ \hline
MahaBERT              & 0.849                                     & 0.849                                     & 0.849                                        \\ \hline
MahaTweetBERT & \textbf{0.854}                                     & \textbf{0.853}                                     & \textbf{0.853}                                        \\ \hline
\end{tabular}

\caption{Results of MahaTweetBERT in comparison with other models on the Marathi Sentiment Analysis dataset, MahaSent/}
\label{tab:mahasent_results}
\end{table*}

\end{document}